\documentclass[final]{cvpr}

\usepackage{times}
\usepackage{epsfig}
\usepackage{graphicx}
\usepackage{amsmath}
\usepackage{amssymb}
\usepackage{bm}
\usepackage{multirow}
\usepackage{stmaryrd}
\usepackage{diagbox}
\usepackage{booktabs} 
\usepackage{makecell}

\usepackage[pagebackref=true,breaklinks=true,colorlinks,bookmarks=false]{hyperref}

\def\cvprPaperID{5460} % *** Enter the CVPR Paper ID here
\def\confYear{CVPR 2021}

\begin{document}

%%%%%%%%% TITLE
\title{Are Labels Always Necessary for Classifier Accuracy Evaluation?} 
\author{Weijian Deng~ ~  Liang Zheng\\
	 Australian National University \\
   {\tt\small ~~~\{firstname.lastname\}@anu.edu.au}
}
%******************
\maketitle

\maketitle
%\thispagestyle{empty}

%%%%%%%%% ABSTRACT
\begin{abstract}
   To calculate the model accuracy on a computer vision task, e.g., object recognition, we usually require a test set composing of test samples and their ground truth labels. 
Whilst standard usage cases satisfy this requirement, many real-world scenarios involve unlabeled test data, rendering common model evaluation methods infeasible. We investigate this important and under-explored problem, Automatic model Evaluation (AutoEval). Specifically, given a labeled training set and a classifier, we aim to estimate the classification accuracy on unlabeled test datasets. We construct a meta-dataset: a dataset comprised of datasets generated from the original images via various transformations such as rotation, background substitution, foreground scaling, etc. As the classification accuracy of the model on each sample (dataset) is known from the original dataset labels, our task can be solved via regression. Using the feature statistics to represent the distribution of a sample dataset, we can train regression models (e.g., a regression neural network) to predict model performance. Using synthetic meta-dataset and real-world datasets in training and testing, respectively, we report a reasonable and promising prediction of the model accuracy. We also provide insights into the application scope, limitation, and potential future direction of AutoEval.
\end{abstract}

%%%%%%%%% BODY TEXT
\section{Introduction}

\begin{figure*}
\begin{center}
\includegraphics[width=0.95\linewidth]{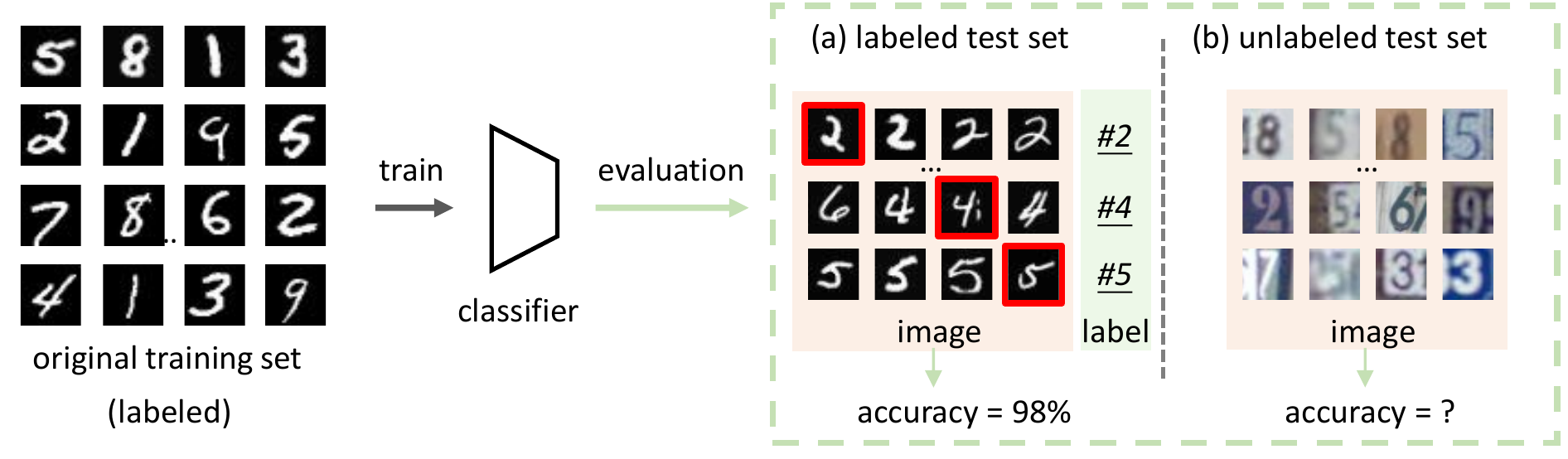}
\caption{Problem illustration. Given a classifier trained on a training set, we can gain a hopefully unbiased estimate of its real-world performance by evaluating it on an unseen labeled test dataset, as shown in (a). However, in many real-world deployment scenarios, we are presented with unlabeled test datasets (b), and as such are unable to evaluate our classifier using common metrics. This inspired us to explore the problem of Automatic model Evaluation.}
\end{center}
\vspace{0.5em}
\label{figure 1}
\end{figure*}

Model evaluation is an indispensable step in almost every computer vision task. Using a test set that is unseen during training, the goal of evaluation is to estimate a model's (hopefully) unbiased accuracy when deployed in real-world scenarios. In most cases, we are provided with a labeled test set, allowing us to calculate the accuracy of a model by comparing the predicted labels with the ground truth labels (\emph{e.g.}, Fig. \ref{figure 1}(a)). In the community, there are many well-established benchmarks (\eg, ImageNet \cite{deng2009imagenet} and COCO \cite{lin2014microsoft}) that provide various types of evaluation metrics. For example, top-1 error, commonly used in image classification, indicates whether the predicted class is the same as the ground truth. There are some other metrics such as mean average precision in object detection \cite{lin2014microsoft} and panoptic quality \cite{kirillov2019panoptic} in panoptic segmentation.

Compared with the evaluation on these benchmarks, evaluating model performance for real-world deployment is not that straightforward. Often, real-world data follow distributions that differ from the original training distribution. In this case, a model's performance on the test sets in a benchmark may not reflect that achieved during deployment. If we still need to have an estimation of the model's accuracy in this scenario, we have to re-evaluate it on the real-world data. However, we often face scenarios where annotations of test samples are not provided. Furthermore, it can be very complex and expensive to manually gather labels. Even if acquired, these samples may only cover a very limited set of conditions, adding bias to the evaluated performance.  
For example, it is very expensive to annotate test samples for license plate recognition systems; even label is gathered for every car, it still can not capture the diversity of real-world circumstances such as lighting and weather condition. This raises an interesting question: \emph{can we estimate model performance on a test set without test labels?}

To answer this question, this paper introduces the Automatic model Evaluation (AutoEval) problem.
Given a classifier trained on a training set, the goal is to estimate its accuracy on an unlabeled test set. 
Here, we introduce an example in Fig. \ref{figure 1}(b). 
Given a digit classifier trained on MNIST \cite{lecun1998gradient}, we want to predict the classification accuracy on a test set \emph{without} ground truths.  
This problem is challenging, as a test set contains many images, and each image has varied and rich visual contents.
However, by visually inspecting the obvious differences between test and training sets, we can infer that the accuracy on the test set is low.

%%%%%%%%%%%%%%%%%%%%%%%%%%%%%%%%%%%%%%%%%%%%%%%%%%%%%%%%%%%%%
\setlength{\tabcolsep}{2.0pt}
\begin{table}
\begin{center}
\begin{tabular}{c|c|c}
	\Xhline{1.2pt} 	
    
                     & Image Classification                                                    & AutoEval                                                                                        \\ \hline
    Sample           & Image                                                                   & Dataset (sample set)                                                                            \\ \hline
    Label            & \begin{tabular}[c]{@{}c@{}}Sample class \\ ground truth \end{tabular} & \begin{tabular}[c]{@{}c@{}}Accuracy of model \\ on sample set \end{tabular}                                    \\ \hline
    Train Set & \begin{tabular}[c]{@{}c@{}} Set of \\ labeled images   \end{tabular}                                                 & \begin{tabular}[c]{@{}c@{}}Set of synthetic labeled \\ sample sets (meta set)\end{tabular}               \\ \hline
    Test Set  & \begin{tabular}[c]{@{}c@{}}Set of unseen \\ labeled images\end{tabular} & \begin{tabular}[c]{@{}c@{}}Set of unseen \\ labeled real-world datasets\end{tabular}            \\ \hline
    Loss             & Class cross-entropy                                                     & Predicted accuracy RMSE                                                                           \\ \hline
    Task             & Classify images                                                         & \begin{tabular}[c]{@{}c@{}}Predict accuracy of model \\ from statistics of dataset\end{tabular}  \\ 	\Xhline{1.2pt} 	
\end{tabular}\label{analogy}
\end{center}
\caption{Analogies between standard image classification terms and their AutoEval equivalents. The analogy shows that the image classification is an image-based task, while the AutoEval problem in this work is dataset-based.
}
\end{table}
%%%%%%%%%%%%%%%%%%%%%%%%%%%%%%%%%%%%%%%%%%%%%%%%%%%%%%%%%%%%%
From this observation, we study AutoEval by considering the distribution difference between training and test sets and how it effects classifier accuracy. Existing literature gives us important hints. Data distributions can be represented by first and second-order statistics of the mean vector of output image feature representations \cite{sun2016return,peng2019moment,GrettonBRSS06}. For example, distribution difference can be estimated via Fréchet Distance (FD) \cite{dowson1982frechet} or maximum mean discrepancy (MMD) metric \cite{GrettonBRSS06}. In addition, domain adaptation literature shows that a smaller distribution difference leads to higher target domain accuracy and implies that a large domain gap causes a low test accuracy \cite{GaninL15,TzengHSD17,TzengHZSD14}. 

In this work, we explicitly show that there is a very strong negative correlation  between accuracy and distribution difference (the Spearman's Rank Correlation \cite{spearman1961proof} is $-0.9$). This observation indicates that it is feasible to estimate classifier accuracy with distribution statistics.
With this, we attempt to quantitatively estimate the test accuracy by studying the underlying relationship between dataset distribution and classifier performance. 
We propose to learn this relationship via a meta-dataset (dataset of datasets). We use the terms meta set and meta-dataset interchangeably. Unlike most existing datasets that treat each image as a sample, we focus on the dataset level: in the meta-dataset, each dataset is treated as a sample, which we term ``sample set''. The analogy between standard image classification and AutoEval task is shown in Table \ref{analogy}.
The sample sets should possess an appropriate number of images, exhibit a diverse spread of distributions, and in the case of image classification, have the same set of classes. 

It is difficult to collect sufficient real-world sample sets that meet the above mentioned requirements. In this work, we propose to construct the meta set by data synthesis.
Every sample set in the meta set is generated from a seed set that follows the same distribution as the original training set. This is achieved via various geometric and photo-metric transformation operations on the seed set, including blurring, background substitution, foreground rotation, translation, \emph{etc}. Note that, the synthetic sample sets are fully labeled because they are transformed versions of the seed set. Using these labels, we can obtain the recognition accuracy of the classifier on each sample set. Sample set $i$ can thus be denoted by $({\bm f}_i, a_i)$, where $a_i$ is a recognition accuracy, and ${\bm f}_i$ is the vector representation of the dataset, \emph{e.g.,} the mean vector of image features in this dataset. With this meta set denoted as $\{({\bm f}_i, a_i)\},i=1,...,N$, where $N$ is the number of sample sets, we can train a regression model that takes input as the  ${\bm f}$ of a sample set and returns the predicted classifier accuracy on this set. 

In summary, the main contributions of this paper include:
\begin{itemize}
    \item We introduce the AutoEval task, aiming to estimate the recognition accuracy of a trained classifier on a test set \emph{without} any human annotated label. 
    \item We validate the feasibility of estimating classifier accuracy from dataset-level feature statistics. With this, we propose to learn an accuracy regression model from a synthetic meta-dataset (a dataset comprised of many datasets) and obtain promising accuracy predictions for real-world test datasets. 
\end{itemize}

\section{Automatic Model Evaluation}

We are interested in predicting the recognition accuracy of a trained classifier on an unlabeled test set. 

%-------------------------------------------------------------------------
\subsection{Problem Definition}

We first define a labeled dataset, $D^l=\{(\bm{x}_i, y_i)\}$ where $i \in [1,...,M]$,  $\bm{x}_i$ is an image, $y_i$ is its class label, and $M$ is the number of images. Consider a source domain $\mathcal{S}$, from which we sample an original training dataset $\mathcal{D}_{ori}$.
We use $\mathcal{D}_{ori}$ to train a classifier $f_{\bm{\theta}}: \bm{x_i}\rightarrow	\hat{y_i}$, which is parameterized by $\bm{\theta}$ and maps an image $\bm{x_i}$ to its predicted class $\hat{y_i}$. 
Given $D^l$, we obtain its classification accuracy by comparing the class predictions $\hat{y_i}$ with the ground truth $y_i$ to obtain accuracy, %we calculate accuracy of $f_{\theta}$ on a labeled test set  $\mathcal{D}_t=\{(\bm{x}_i, y_i), i=1,...,N\}$ of $N$ images by comparing predictions $\hat{y_i}$ with ground truths $y_i$,
\begin{equation}
    a_{standard} = \frac{\sum_{i=1}^M\llbracket{\hat{y_i}==y_i}\rrbracket}{M},
\end{equation}
where $\llbracket{\cdot\rrbracket}$ is an indicator function returning $1$ if argument is true and $0$ otherwise.

In AutoEval, given $f_{\bm{\theta}}$ and an unlabeled dataset {$\mathcal{D}^u=\{\bm{x}_i$\}} for $i \in [1,...,M]$, we use an accuracy predictor $A: (f_{\bm{\theta}}, \mathcal{D}^u) \rightarrow a$, which outputs an estimated classifier accuracy $a \in [0,1]$ on this test set,
\begin{equation}
    a_{auto} = A(f_{\bm{\theta}}, \mathcal{D}^u).
\end{equation}
Note that in image classification, $\mathcal{D}_{ori}$ and $\mathcal{D}^u$ share the same label space.

%-------------------------------------------------------------------------
\subsection{An Intuitive Solution}\label{sec:confidence}

We first present an intuitive solution to the AutoEval problem, which is not learning based. This solution is motivated by the pseudo labeling strategy in many vision tasks \cite{hendrycks2016baseline,zhang2018collaborative,ma2017self}. The basic assumption is: if a class prediction is made with a high softmax output score, this prediction is likely to be correct. Formally, let us consider a $K$-way classification problem. When feeding a test image $x_i$ to a trained classifier $f_{\bm{\theta}}$, we obtain $\bm{s}_i\in \mathbb{R}^K$, which is the output of the softmax layer. The $k$-th entry in $\bm{s}_i$ characterizes the probability of $x_i$ belonging to class $k$. The $\ell_1$ norm $\|\bm{s}_i\|_1=1$. 
If the maximum entry of $\bm{s}_i$ is greater than a threshold $\tau$, image $x_i$ is considered to be correctly classified. The accuracy predictor is written as,
\begin{equation}
    a_{max} = A_{max}(f_{\bm{\theta}},\mathcal{D}^u)=\frac{\sum_{i=1}^M\llbracket{\max(\bm{s}_i)>\tau}\rrbracket}{M},
\label{eq:max}
\end{equation}
where $M$ is the number of images in $\mathcal{D}^u$. We will evaluate $A_{max}$ in the experiment and show that it does not work consistently well across datasets.

%------------------------------------------------------------------------
\section{Methods}

%-------------------------------------------------------------------------
\subsection{Formulation}
Motivated by the implications in domain adaptation, we propose to address AutoEval by measuring the distribution difference between the original training set and the test set, and explicitly learning a mapping function from the distribution shift to the classifier accuracy. 

Under this consideration, we formulate AutoEval as a dataset-level regression problem. In this problem, we view a dataset as a sample, and its label is the recognition accuracy on the dataset itself. Given $N$ sample sets, we denote the $j$-th sample set $\mathcal{D}_j$ as $(\bm{f}_{j}, a_{j})$, where $\bm{f}_{j}$ is the representations for $\mathcal{D}_j$, and $a_j\in [0,1]$ is the recognition accuracy of classifier $f_{\bm{\theta}}$ on $\mathcal{D}_j$. 
We aim to learn a regression model (accuracy predictor), written as,
\begin{equation}
   a_j = A(\bm{f}_j).
\label{eq:regression}
\end{equation}
We use a standard squared loss function for this model, 
\begin{equation}
    \mathcal{L} = \frac{1}{N}\sum_{j=1}^{N} (\hat{a}_{j}-a_{j})^{2},
\label{eq:loss}
\end{equation}
where $\hat{a}_{j}$ is the predicted accuracy of the $j$-th sample set $\mathcal{D}_j$, and $a_{j}$ is the ground truth classifier accuracy of $\mathcal{D}_j$. 

During testing, we extract the dataset representation $\bm{f}^u$ for unlabeled test set $\mathcal{D}^u$, and obtain estimated classification accuracy using $a=A(\bm{f}^u)$.

To learn regression models defined in Eq. \ref{eq:regression} and Eq. \ref{eq:loss}, we need to specify the design of 1) dataset representation $\bm{f}_j$, 2) regression model $A$, and 3) $N$ sample sets (meta-dataset). 

\subsection{Regression Model and Dataset Representation}\label{sec:regression}
\noindent \textbf{Linear regression.} We first introduce a simple linear regression model, 
\begin{equation}\label{linear}
    a_{linear} = A_{linear}(\bm{f}) = w_1 f_{linear}+w_0,
\end{equation}
where $f_{linear}\in \mathbb{R}$ is the representation of sample set $\mathcal{D}$, and $w_0, w_1 \in \mathbb{R}$ are parameters of this linear regression model. 
Based on the intuition that the domain gap impacts classifier accuracy, we define $f_{linear}$ as the quantified domain gap between dataset $\mathcal{D}$ and the original training set $\mathcal{D}_{ori}$. Specifically, we use the Fréchet distance \cite{dowson1982frechet} to measure the domain gap, and thus,
\begin{equation}
\begin{split}
f_{linear} = \mbox{FD}(\mathcal{D}_{ori}, \mathcal{D}) &=  \left \| \bm{\mu}_{ori} - \bm{\mu}  \right \|^{2}_{2} + \\ & Tr(\bm{\Sigma}_{ori} + \bm{\Sigma} -2 (\bm{\Sigma}_{ori} \bm{\Sigma})^{\frac{1}{2}}),
\end{split}
\label{eq:fd}
\end{equation}
where $\bm{\mu}_{ori}$ and $\bm{\mu}$ are the mean feature vectors of $\mathcal{D}_{ori}$ and $\mathcal{D}$, respectively. $\bm{\Sigma}_{ori}$ and $\bm{\Sigma}$ are the covariance matrices of $\mathcal{D}_{ori}$ and $\mathcal{D}$, respectively. They are calculated from the image features in $\mathcal{D}_{ori}$ and $\mathcal{D}$, which are extracted using the classifier $f_{\bm{\theta}}$ trained on $\mathcal{D}_{ori}$. 
Other measurements of the domain gap can also be used, such as MMD  \cite{GrettonBRSS06}. 

%=================================FIG 2 ======================%
 \begin{figure*}[t]
 \begin{center}
\includegraphics[width=0.98\linewidth]{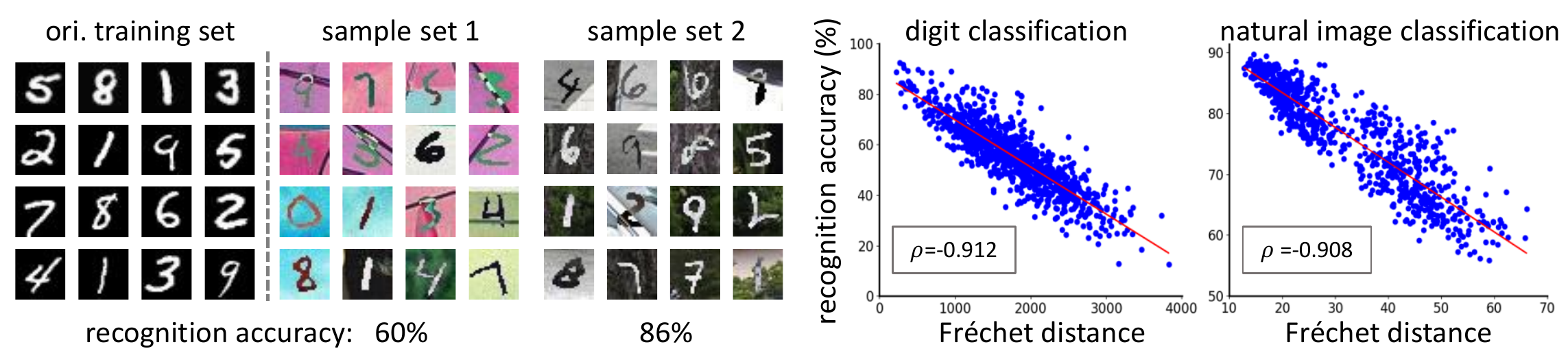}
\caption{Relationship between the distribution shift and classifier accuracy on digits and natural image classification. Each point represents a sample set of the meta set. 
The Spearman's Rank Correlation ($\rho$) \cite{spearman1961proof} between distribution shift and classifier accuracy is around $-0.91$ on two scenarios, indicating they have a very strong negative correlation. The red straight line is fit with robust linear regression \cite{huber2004robust}.}
\label{fig:fid}
 \end{center}
\end{figure*}
%=================================FIG 2 ======================%

\paragraph{Proof of concept.}
Given a meta set and a classifier trained on the training dataset $\mathcal{D}_{ori}$ from a source domain $\mathcal{S}$, we study the relationship between classifier's accuracy and distribution shift. 
In Fig. \ref{fig:fid}, we show the accuracy as a function of the distribution shift. 
The distribution shift is measured by Fréchet distance (FD) with the features extracted from the trained classifier.
In practice, we use the activations in the penultimate of the classifier as features.

In both digits and natural image classification, we observe a very strong negative correlation between accuracy and distribution shift in both digits and natural image classification: the Spearman's Rank Correlation  ($\rho$) \cite{spearman1961proof} is about $-0.91$. 
That is, the classifier tends to achieve a low accuracy on the sample set which has a high distribution shift with training set $\mathcal{D}_{ori}$. 
This indicates it is feasible to learn a regression model to predict classifier performance based on distribution difference between training and test sets.

\noindent\textbf{Neural network regression.} Besides the linear regression, we also propose a neural network regression model, $a_{neural} = A_{neural}(\bm{f}_{neural})$, which has the same formulation as Eq. \ref{eq:regression}. 
In practice, we use a simple fully connected neural network for regression.
The input of the model is the dataset representation $\bm{f}_{neural}$, and the output is the estimated classifier accuracy $a_{neural}$.

With the observation in the proof of concept, we propose to use distribution-related statistics to represent a dataset. In this work, we use its first-order and second-order feature statistics, \emph{i.e.}, mean vector and covariance matrix.
Moreover, we also include a 1-dim FD score as an auxiliary information to the representation. 
Compared with linear regression, the neural network regression has a richer dataset representation.
The dataset representation is written as, 
\begin{equation}
\begin{split}
\bm{f}_{neural} = [f_{linear}; \bm{\mu}; \bm{\sigma}],
\end{split}
\label{eq:neural}
\end{equation}
where $f_{linear}\in \mathbb{R}$ is the Fréchet distance between $\mathcal{D}$ and $\mathcal{D}_{ori}$, $\bm{\mu}$ and $\bm{\Sigma}$ are calculate the same way as Eq. \ref{eq:fd}. 
Covariance $\bm{\Sigma} \in \mathbb R^{d\times d}$ is very high-dimensional, making training difficult. Dimension reduction is thus necessary. Specifically, we calculate $\bm{\sigma}$ by taking a weighted summation of each row of $\bm{\Sigma}$ to produce a single vector, using learned column specific coefficients that are shared across all rows. For example, if the feature extracted from $f_{\bm{\theta}}$ is $d$-dim, the dimensionality of $\bm{f}_{neural}$ is $1+ 2d$.

% --------------------------------------------- FIG 2 -----------------------------------------#
 \begin{figure}[t]
 \begin{center}
\includegraphics[width=0.98\linewidth]{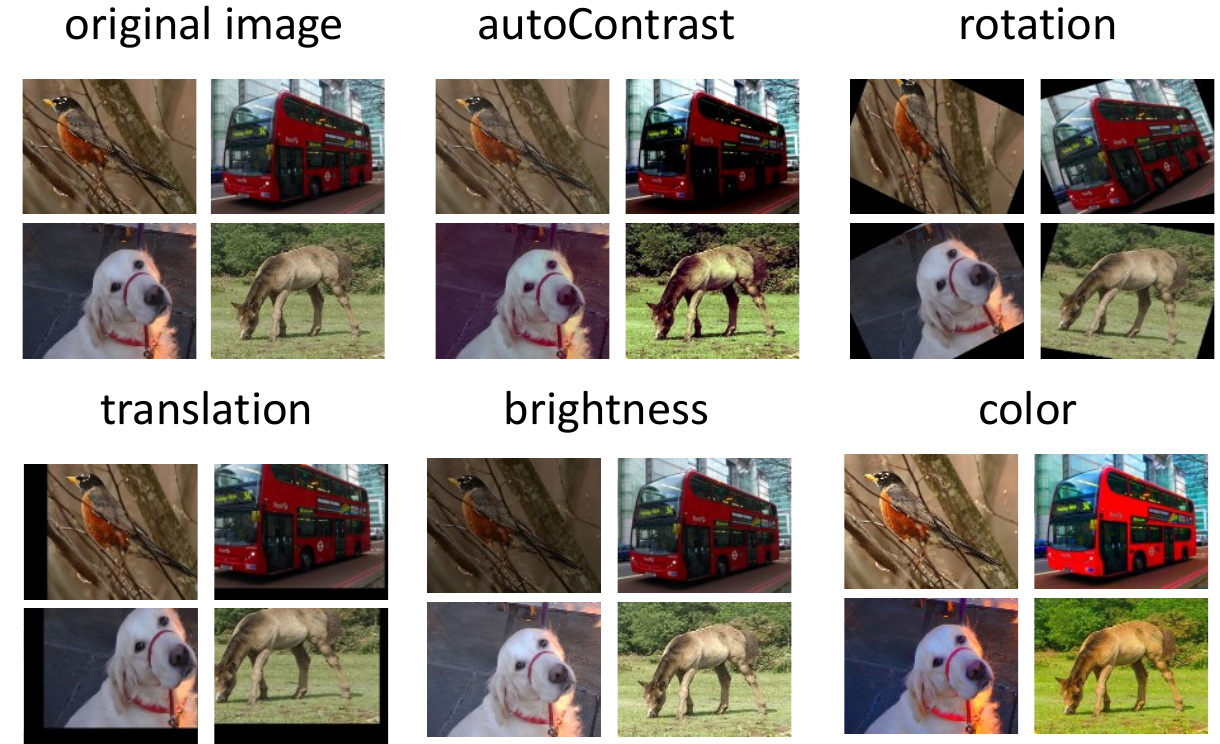}
\caption{Visual examples of transformations. Here we show autoContrast, rotation, translation, brightness, and color. For other used transformations, we refer readers to \cite{cubuk2019autoaugment}.}\label{fig:transformations}
 \end{center}
\end{figure}

% --------------------------------------------- FIG 2 -----------------------------------------#
% --------------------------------------------- FIG 3 -----------------------------------------#
 \begin{figure}[t]
 \begin{center}
\includegraphics[width=1.0\linewidth]{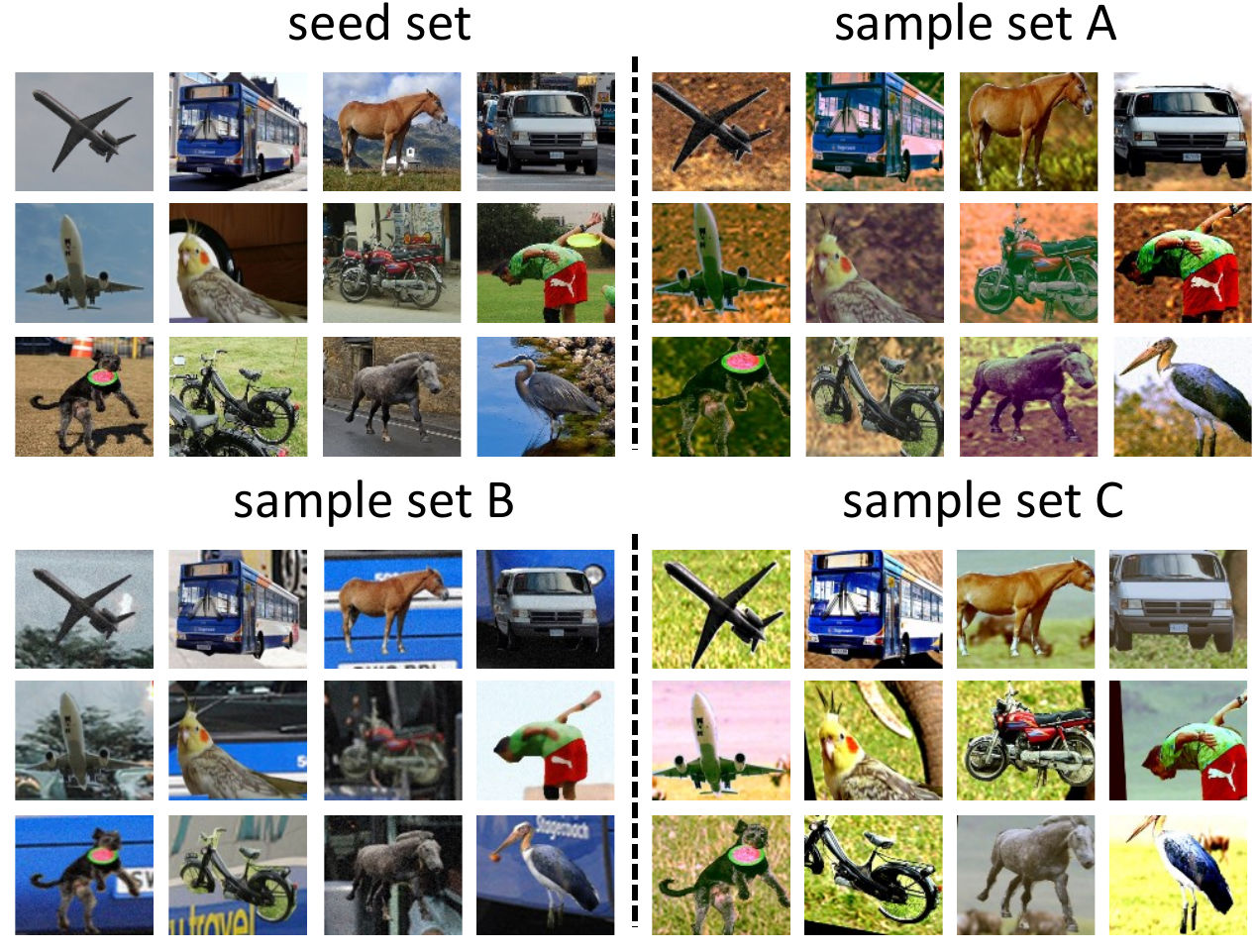}
\caption{The seed set and examples of three sample sets. The seed set is from the same distribution with the original training set; they share the same classes but do not have image overlap. The sample sets are generated from the seed by background replacement and image transformations. The sample sets exhibit distinct data distributions, but inherit the foreground objects from the seed, and thus are fully labeled. Many sample sets form a meta-dataset from which an accuracy regression model is trained.}
\label{fig:sampleset}
\end{center}
\end{figure}

% -------------------------- FIG 3 -----------------------------#

\subsection{Constructing Training Meta-dataset}
\noindent\textbf{Meta-datasets for training.} The regression model (Eq. \ref{eq:regression}, Eq. \ref{eq:loss}, Eq. \ref{eq:neural}) takes the dataset representation as input and outputs a classification accuracy. To train it, we need to prepare a meta-dataset in which each sample is a dataset. 
In classification, the diversity of the samples in the training set should ideally be sufficient such that test scenario is represented in its distribution. In this work, we seek to create a diverse meta set that (hopefully) contains the test distributions.
To construct such a meta set, we should collect sample sets that are 1) large in number, 2) diverse in the data distribution, and 3) have the same label space with the training set.
There are very few real-world datasets that satisfy these requirements, so we resort to data synthesis. 

For each classification task (digits or natural images), we synthesize sample sets from a single seed dataset. The seed $\mathcal{D}_s$ is sampled from source domain $\mathcal{S}$, and thus has the same distribution as $\mathcal{D}_{ori}$. Given $\mathcal{D}_s$, we apply various visual transformation and obtain $N$ different sample sets $\mathcal{D}_j, j=1,...,N$. Since $\mathcal{D}_s$ is fully labeled, these sample sets inherent the labels from $\mathcal{D}_s$. 

To create a sample set $\mathcal{D}_j$, we adopt a two-step procedure: perform background change, and then image transformations.
In the first step, we keep the foreground / object unchanged and replace the background. For each sample set, we randomly select an image from the COCO dataset \cite{lin2014microsoft}, from which we randomly crop a patch and use it as the background. The patch scale and position in that image are both random. In the second step, for the background-replaced images, we use six image transformations defined in \cite{cubuk2019autoaugment}, including autoContrast, rotation, color, brightness, sharpness, and translation.
Examples of some transformations are shown in Fig. \ref{fig:transformations}. For each sample set, we randomly select and combine three out of the six transformations, with the magnitude of each transformation being random on per-sample basis. As such, each sample set is generated by background replacement and a combination of three image transformations.
Fig. \ref{fig:sampleset} presents examples of sample sets in natural image classification, where background replacement can be observed. In the supplementary materials, we present the detailed transformation parameters and more visual examples of the training meta set. 
Note that a sample set inherits all the image labels from the seed set and is fully labeled. As such, we can calculate the recognition accuracy of classifier $f_{\bm{\theta}}$ on each sample set. Sample set $\mathcal{D}_j$ can be denoted as $(\bm{f}_j, a_j)$, which is used as a training sample to optimize the regression model. % is transformed from the seed set

\noindent\textbf{Real-world datasets for testing.} 
This is an early attempt for the AutoEval problem. To our knowledge, we could only find few real-world datasets that have different distributions but contain the same classes. To clarify the AutoEval problem, we conduct extensive analyses with these dataset. For digits classification, we use USPS \cite{hull1994database} and SVHN \cite{netzer2011reading}, both with 10 classes. For natural image classification, we use three existing datasets, \emph{i.e.,} PASCAL \cite{everingham2007pascal}, Caltech \cite{griffinHolubPerona}, and ImageNet \cite{deng2009imagenet}, all with 12 classes. 
Details of the test meta sets are provided in Sec. \ref{sec:settings}.
%================================Table 2 ===========================%
\setlength{\tabcolsep}{13pt}
\begin{table*}[t]
\small
    \begin{center}
    \begin{tabular}{l|cc|c|ccc|c}
 \Xhline{1.2pt} 	
    Train Set&\multicolumn{3}{c|}{Digits}&\multicolumn{4}{c}{Natural images}\\
    \cline{1-8}
     Unseen Test Set & SVHN &  USPS & RMSE$\downarrow$ & Pascal & Caltech & ImageNet & RMSE$\downarrow$\\
    \hline
    Ground-truth accuracy & 25.46& 64.08 & - & 86.13&93.40&88.83&-\\
    \hline
    Predicted score ($\tau =0.7$) & 10.09& 43.60&18.11 &88.34&93.28&90.17&\boxed{1.49}\\
    Predicted score ($\tau =0.8$) & 7.97& 37.22&22.66 &84.32&90.78&86.50&{2.28}\\
    Predicted score ($\tau =0.9$) & 7.03& 32.94&25.59 &78.61&87.71&81.33&6.96 \\
    \hline
    \hline
    Linear reg.  & 26.28 & 50.14&9.87&83.87 &79.77 & 83.19&8.62\\
    Neural network reg.  & 27.52 &64.11 &\boxed{1.46}&87.76& 89.39& 91.82&\boxed{3.04}\\
\Xhline{1.2pt} 	
    \end{tabular}
    \end{center}
    \caption{Method comparison in predicting classification accuracy. Results on digit classification (SVHN and USPS datasets) and natural image classification (Pascal, Caltech, and ImageNet) are shown. We compare three methods, \emph{i.e.,} predicted score based (Section \ref{sec:confidence}), linear regression and neural network regression (Section \ref{sec:regression}). For each dataset, we report the estimated classification accuracy (\%). For both digit and natural image classification, RMSE (\%) is reported. The original training sets are MNIST and COCO, respectively. The ground-truth recognition accuracy (\%) is presented.} 
    \label{tab:digits}
\end{table*}
%%%%%%%%%%%%%%%%%%%%%%%%%%%%%%%%%%%%%%%%%%%%%%%%%%%%%%%%%%%%
\section{Experiment and Analysis}
\subsection{Experimental Settings}\label{sec:settings}
We study the AutoEval problem on two classification tasks: digit classification and natural image classification. 

\noindent\textbf{Digit classification.} The original training set contains all the training images of MNIST. We use the testing images of MNIST as the seed to generate the training meta set. 
Because MNIST images are binary, the foreground can be separated from the background. When generating a meta set, we randomly select an image from the COCO training set, and the background of each image is replaced with a random patch of the sampled COCO image.
Then, we apply three out of six image transformations to images. We generate $3,000$ sample sets, of which we use 2,000 and 1,000 for the training and the validation meta set, respectively. 
Moreover, we use two real datasets for testing, \emph{i.e.,} USPS \cite{hull1994database} and SVHN \cite{netzer2011reading} datasets. 

 \begin{figure}[t]
 \begin{center}
\includegraphics[width=1.0\linewidth]{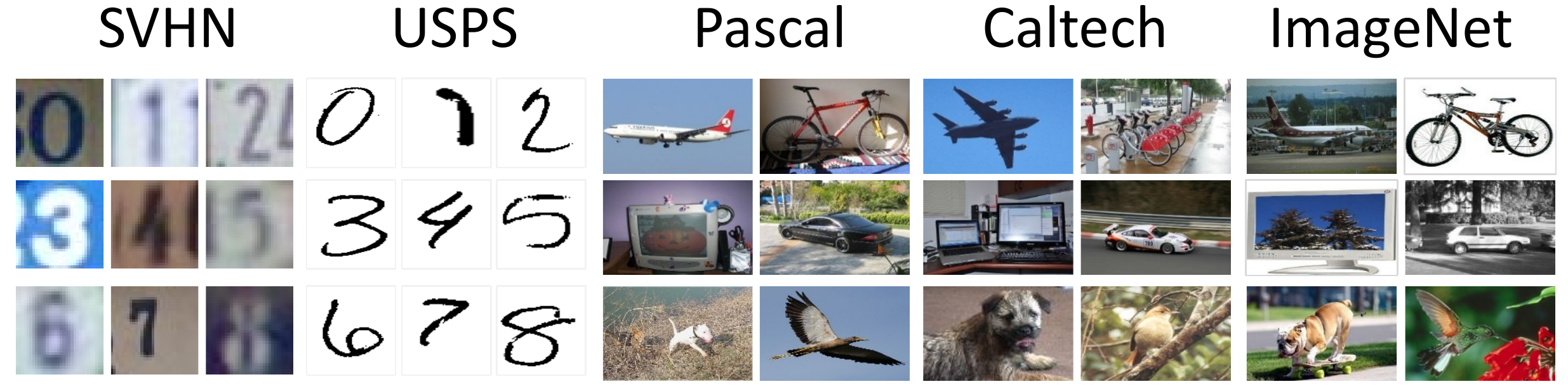}
\caption{Sample images from real-world test datasets, including SVNH, USPS, Pascal, Caltech and ImageNet. The former two are for digit classification, and the latter three are for natural image classification. We predict the classifier accuracy on these datasets.}
\label{fig:testdata}
\end{center}
\end{figure}

\noindent\textbf{Natural image classification.} We use COCO \cite{lin2014microsoft} training set as the original training set, and COCO validation set as the seed set to build meta set. When generating meta set for training, we use instance mask annotations of COCO validation set to get foreground regions. Similar to digit classification, for each sample set, we replace the background with a random patch of an image from COCO test set. We then use image transformations to introduce more visual changes. We create 1,600 sample sets from the seed set, of which we use 1,000 and 600 for the training and the validation meta set, respectively. In testing, we use PASCAL \cite{everingham2007pascal}, Caltech \cite{griffinHolubPerona}, and ImageNet \cite{deng2009imagenet}. For each dataset, we select images of 12 common classes, \emph{i.e.}, aeroplane, bike, bird, boat, bottle, bus, car, dog, horse, monitor, motorbike, and person. We reduce the ``person'' class to 600 images to balance the overall number of images per class.

\noindent\textbf{Classifier architecture.} For digit classification, we use LeNet-5 \cite{lecun1998gradient} as classifier.  %In the experiment, the digits classifier is based on  and is learned on the original training set. 
Since the all images are mapped to the RGB space, we modify the number of input channel of LeNet-5 to 3. For natural image classification, we use the ResNet-50 pretrained on ImageNet \cite{deng2009imagenet} which is adapted to the 12-way classification.

\noindent\textbf{Metrics.} This paper estimates the recognition accuracy of a model on a test set. To evaluate the performance of such estimate, we use root mean squared error (RMSE) and mean absolute error (MAE) as metrics. 
RMSE measures the average squared difference between the estimated classifier accuracy and ground-truth accuracy. MAE measures the average magnitude of the errors. Small RMSE and MAE correspond to good predictions and vice versa. 

%================================Table 2 ===========================%

\subsection{Classifier Accuracy Prediction}
This paper introduces three possible methods to estimate the recognition accuracy, including the confidence-based method, linear regression and neural network regression. We report the estimations of these methods in Table \ref{tab:digits}. For the predicted score based method, three thresholds(\emph{i.e.}, $\tau=0.7$, $0.8$and $0.9$ in Eq. \ref{eq:max}) are used. 

\noindent\textbf{The  predicted score based method is very sensitive to the threshold.} Under a specific threshold ($\tau=0.7$), this method makes accuracy prediction on natural image datasets (RMSE=1.49\%), but its prediction quality drops significantly (from $1.49\%$ to $6.96\%$) when we increase value of $\tau$ to $0.9$. What is more, its performance is very poor when considering the digit classification task. Under two values of $\tau$, the RMSE is consistently high, \emph{i.e.,} 22.66\% and 25.59\%, respectively. 
Note that, it is infeasible to select the optimal threshold because 1) test labels are unavailable and 2) the test domain keeps changing. Our method does not depend on such a hyper-parameter and yields much more stable results.
That said, it would be interesting to address this drawback in the context of AutoEval.

\noindent\textbf{Regression methods achieve better predictions than predicted score based method.} In digit datasets, the RMSE values of linear regression and neural network regression are 9.87\% and 1.46\%, respectively. A similar trend can be observed in natural image datasets. Their RMSE scores are generally lower and more stable than the predicted-based method. This indicates the effectiveness of learning-based methods: the distribution difference between the original training and test sets is a critical feature. 
%==========================FIG6=============================%
\begin{figure}[t]
\begin{center}
\includegraphics[width=1.0\linewidth]{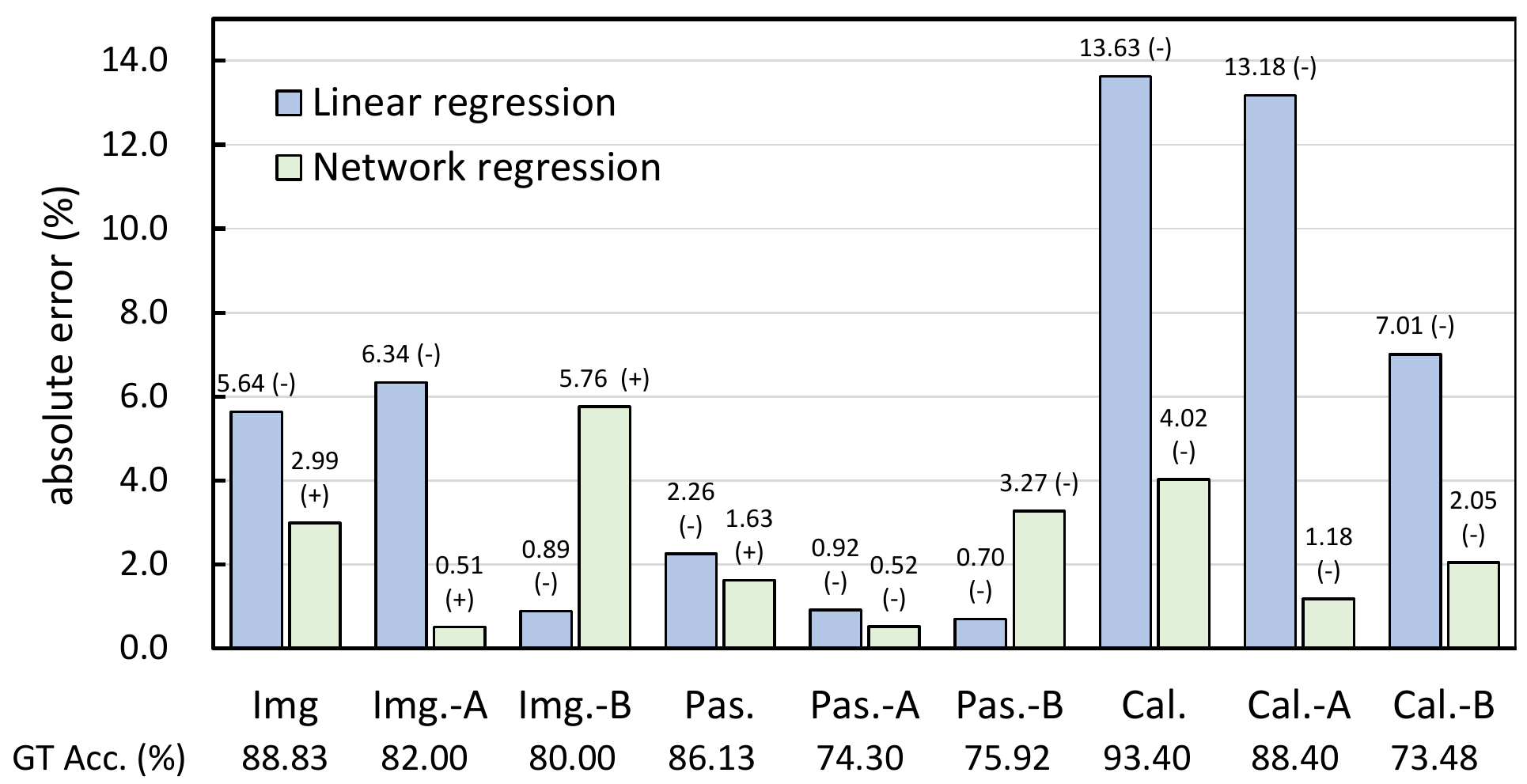}
\caption{Comparing linear regression and neural network regression when test data undergo new image transformations such as Cutout \cite{devries2017improved,zhong2017random}, Shear, Equalize and ColorTemperature \cite{cubuk2019autoaugment}.
The transformed datasets are denoted by ``-A'' and ``-B''.
We report the absolute error (\%) of predictions and the ground truth accuracy is also shown below each dataset.  (-) / (+) means the predicted accuracy is lower / higher than the ground-truth accuracy, respectively.}
\vspace{-2em}
\label{fig:real_transform}
\end{center}
\end{figure}
%==========================FIG6=============================%
%================================ FIG ===========================%
 \begin{figure*}[th]
 \begin{center}
\includegraphics[width=1.0\linewidth]{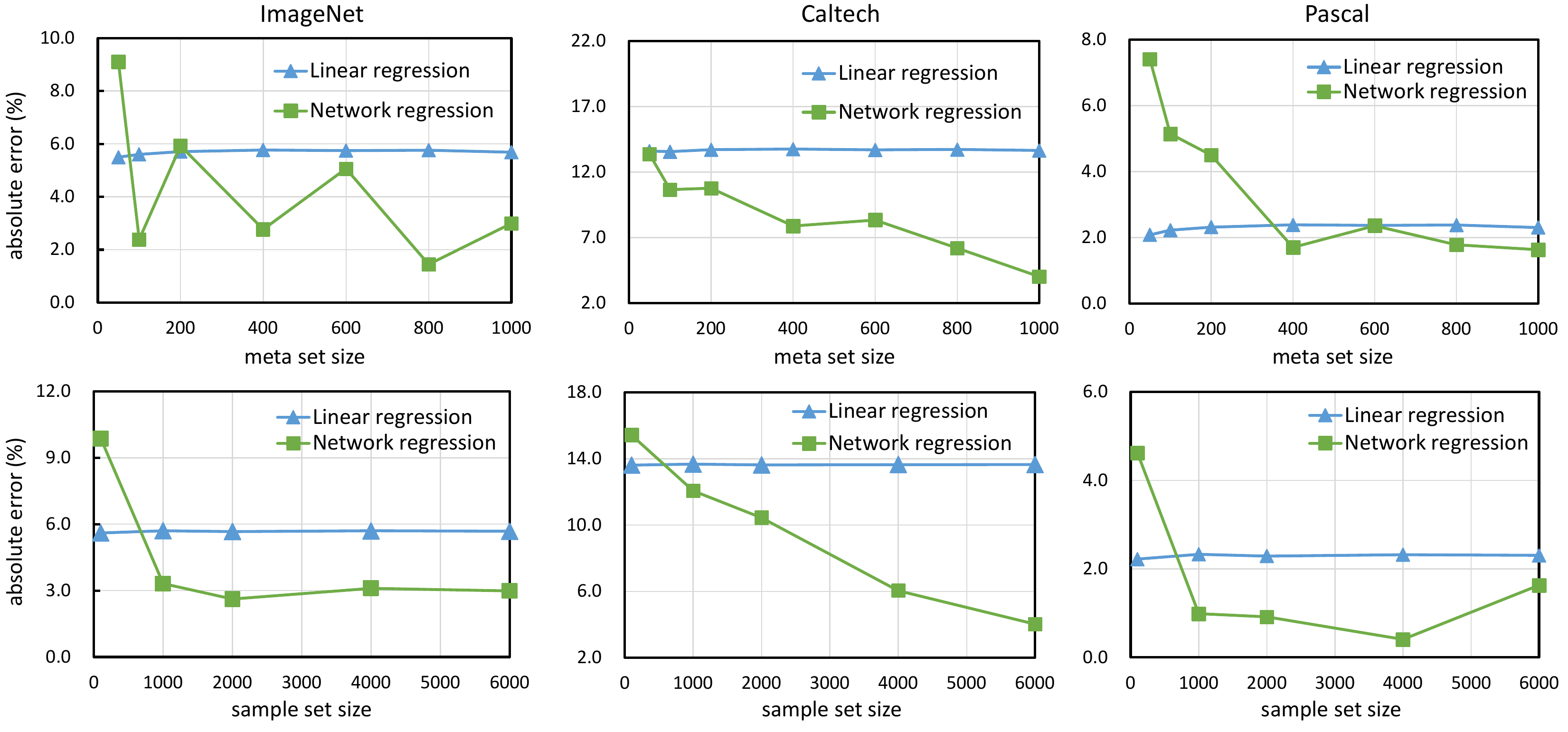}
\caption{The impact of meta set size (first row) and sample set size (second row) on the performance of regression methods. We report the absolute errors (\%) between estimated classifier accuracy and the ground-truth accuracy. We observe that linear regression is relatively stable with different sample set and meta set size. In comparison, neural network needs more and larger sample sets for training.}
\vspace{-1em}
\label{fig:meta-size}
 \end{center}
\end{figure*}
%================================ FIG ===========================%

\noindent\textbf{Neural network regression is generally better than linear regression.} As shown in Table \ref{tab:digits}, the neural network regression is more accurate than linear regression in both digit and natural image datasets. For example, RMSE of the former is 8.41\% lower than the latter on digit datasets. In fact, the RMSE of neural network regression is as small as 1.46\%: the predicted classifier accuracy is very close to the ground truth accuracy.

We note that linear regression is significantly inferior to neural network regression on Caltech datasets, where linear regression gives errors higher than 10\%. 
 Caltech is an interesting dataset. Its images have relatively simple backgrounds and salient foregrounds, implying that they are ``easy'' to classify. However, such simple background contrasts significantly with the original training set (COCO), so the FD score between Caltech and COCO is very large. Only looking at the FD score, linear regression tends to predict low accuracy on Caltech. In comparison, neural network regression considers the data statistics of Caltech, such that it can make more accurate prediction. Furthermore, the meta-dataset might already contain sample sets with such ``simple backgrounds'' (large FD), and high recognition accuracy. Under such circumstances, the network has learned to overrule the large FD and instead resort to the ``simple background'' when making predictions.  

\noindent\textbf{Robustness of regression models.} To further examine the two regression methods, 
%we report a new group of experiment on natural image classification in Fig. \ref{fig:real_transform}. Here, 
we perform image transformations to real datasets (ImageNet, Pascal and Caltach) and assess the performance of the two regression methods on these ``edited real-world datasets''. Note that image transformations we use here are \emph{not} applied in meta-dataset generation. Thus, this experiment assesses some generalization ability of the regression methods. From Fig. \ref{fig:real_transform}, we first observe the ground truth accuracy on the edited datasets is lower than that on the original sets. It suggests that the image transformations are introducing visual differences that hinder the classifier performance. The results show that the two regression methods could also achieve 
reasonably good estimated results. For example, linear regression makes promising predictions on 6 out of 9 datasets (It has the same issue discussed above on the 3 Caltech sets). Our network regression model gives lower errors on all 9 datasets. This suggest that our network can learn from diverse and various sets of the meta-set to make an accurate performance prediction.

\subsection{Analysis of the Training Meta-Dataset}
The synthetic meta-dataset is a key component of our system, allowing us to obtain labeled samples sets in a large scale. We analyze its impact on the regression methods from two aspects, \ie, meta set size and sample set size. 

\noindent\textbf{Meta set size.} We first study the impact of meta set size on the regression methods. Meta set contains training samples/datasets for regression models. In Fig. \ref{fig:meta-size} (first row). We observe the results of linear regression are relatively stable with different meta set sizes. It can achieve good performance even with $50$ sample sets. This is because linear regression only has two parameters (Eq. \ref{linear}), which can be learned with relatively few samples \cite{huber2004robust}. In comparison, neural work cannot achieve good results when the number of sample sets is small.
When provided adequate sample sets, the neural network can learn effectively from rich and diverse sample datasets and surpasses the linear regression.

\noindent\textbf{Sample set size.} By default, the number of images in each sample set is equal to that of seed $\mathcal{D}_s$. We study the impact of sample set size on the regression methods. In the experiment, we set the meta set size 1000, and vary the sample set size. In Fig. \ref{fig:meta-size} (second row), we observe linear regression is stable under different sample set sizes. In comparison, the neural network needs more images in each sample set for training. We think more images in each sample set make the distribute-related representations more accurate. This is beneficial for regression learning of network.

%  Moreover, the neural network regression is generally better than linear regression with different classifiers. We also note that based on the features of VGG-16, the linear regression has a comparable regression accuracy with neural network. 

\section{Related Work}

% \textbf{Dataset-level analysis} 
% While most computer vision tasks can be described as some form of image-level analysis, very limited literature exists that explores dataset-level analysis. 
% % 
% Some work has focused on optimizing dataset construction by selecting a subset of images for pretraining for a specific downstream task  \cite{cui2018large,zhang2018fine,yan2020neural}. 
% Specifically, Cui \emph{et al.} \cite{cui2018large} propose a similarity metric to select relevant classes of a given dataset for network pretraining. In a similar line of work, Zhang \emph{et al.} \cite{zhang2018fine} select task-related images from the auxiliary data for training networks. Recently, Yan \emph{et al.} \cite{yan2020neural} introduce a large-scale search engine to find the most useful transfer learning data for the target task.

% Another avenue of research aims to automatically synthesize a labeled training set for downstream tasks \cite{kar2019meta,ruiz2018learning}. They propose to learn to edit the parameters of graphic engines to minimize the content gap between synthetic training data and real test data. 
% %%
% Our work is related in that we also study inter-dataset relationships, but have a distinct purpose. Specifically, this work aims to estimate classifier performance on an unlabeled dataset from the perspective of the distribution difference.

\noindent\textbf{Model generalization prediction.} 
There are some works develop complexity measurements on {training sets} and {model parameters} to predict generalization error \cite{jiang2018predicting,arora2018stronger,corneanu2020computing,jiang2018predicting,neyshabur2017exploring,unterthiner2020predicting}.
Corneanu \etal \cite{corneanu2020computing} use the persistent topology measures to predict the performance gap between training and testing error, even without the need of any testing samples. Jiang \etal \cite{jiang2018predicting} introduce a measurement of layer-wise margin distributions for generalization ability. Neyshabur \etal \cite{neyshabur2017exploring} develop bounds on the generalization gap based on the product of norms of the weights across layers.
Moreover, the agreement score of several classifies' predictions can be used for estimation \cite{madani2004co,platanios2016estimating,platanios2017estimating,donmez2010unsupervised,jiang2020fantastic}.
Our work differs significantly: we focus on the measuring statistics related to {test sets} for prediction. 

\noindent\textbf{Out-of-distribution (OoD) detection.} This task \cite{devries2018learning,hendrycks2016baseline,lee2018simple,vyas2018out,lee2018training} considers the distribution of test samples. Specifically, this task aims to detect test samples that follow a distribution different from the training distribution. 
This has been studied from different views, such as anomaly detection \cite{andrews2016transfer}, open-set recognition \cite{bendale2015towards}, and rejection \cite{cortes2016learning}. For example, Hendrycks \etal \cite{hendrycks2016baseline} use probabilities output from a softmax classifier as indicator to find out-of-distribution samples.
While this task attempts to detect abnormal testing samples, our work considers the overall statistics of all test samples to predict classifier accuracy.

\noindent\textbf{Unsupervised Domain adaptation.}
Our work also relates to unsupervised domain adaptation, where the goal is to use labeled source samples and unlabeled target samples to learn a model that can generalize well on the target dataset \cite{DAN,TzengHZSD14,zuo2019craves,deng2018similarity}.
%
% Existing approaches attempt to eliminate the shift between the source and target distributions. 
%
Many moment matching schemes have been studied for this task \cite{sun2016return,DAN,TzengHZSD14,peng2019moment,sun2016return,zhang2018aligning}. 
Long \emph{et al.} \cite{DAN} and Tzeng \emph{et al.} \cite{TzengHZSD14} utilize the maximum mean discrepancy (MMD) metric \cite{GrettonBRSS06} to learn a shared feature representation.
% Sun \emph{et al.} \cite{sun2016return} propose to perform domain adaptation by matching the second-order of feature distribution.
% Zhang \emph{et al.} \cite{zhang2018aligning} propose to decrease shift by mapping infinite-dimensional matrices in RKHS. 
% Peng \emph{et al.} \cite{peng2019moment} propose to address multi-source domain adaptation by matching feature moments.
In this work, we study the underlying relationship between the model performance and the distribution shift. By leveraging dataset level statistics, we are able to accurately predict model performance on unlabeled test sets.

\section{Conclusions and Perspectives}

This paper investigates the problem of predicting classifier accuracy on test sets without ground-truth labels. It has the potential to yield significant practical value, such as predicting system failure in unseen real-world environments. Importantly, this task requires us to derive similarities and representations on the dataset level, which is significantly different from common image-level problems. We make some tentative attempts by devising two regression models which directly estimate classifier accuracy based on overall distribution statistics. We build a dataset of datasets (meta-dataset) to train the regression model. We show that the synthetic meta-dataset can cover a good range of data distributions and benefit AutoEval on real-world test sets. For the remainder of this section, we discuss the limitations, potential, and interesting aspects of AutoEval.

\noindent\textbf{Application scope.} Our system assumes that variations in the real-world cases can be approximated by the image transformations in the training meta set. With various and diverse sample sets, our system learns to make promising predictions for novel environments. However, if the test datasets exhibit some very special patterns or conditions, the system might not be able to work. 
An example is that the test dataset has an entirely different set of classes, and this test distribution cannot be approximated by the meta-dataset in our work. Under this circumstance, our trained models will still give an estimated accuracy, which is clearly incorrect. 
On a related extreme case, the test dataset might only contain ambiguous and adversarial samples, meaning that the test accuracy could be as poor as random. Such cases are not included in meta-dataset, either. Potentially, the above two issues could be alleviated by including such cases into the meta-dataset with a specific dataset design. Another option is to use out-of-distribution detection techniques to help detect and reject such cases.

\noindent\textbf{Dataset Representation.} Our work relates to an interesting research problem: how to represent a dataset? This problem is more challenging than describing a single image because a dataset contains much more information. 
This work uses distribution-related feature statistics (mean and covariance) to characterize a classification dataset.
We believe there are other potential representations for better representing a dataset. On the other hand, it would be interesting to study the representation in other tasks (\eg, object detection and semantic segmentation), where global feature statistics might not be suitable to characterize a dataset.

\noindent\textbf{Similarities between datasets.} We measure dataset similarity using the FD score. However, this problem is as challenging as dataset representation, especially when we aim to connect the similarity with test accuracy. This problem will benefit the domain adaptation field, where more precise domain bias measurement and its connection to target set accuracy will significantly help algorithm design. 

\section*{Acknowledge}
This work was supported by the ARC Discovery Early Career Researcher Award (DE200101283) and the ARC Discovery Project (DP210102801). We thank all anonymous reviewers and ACs for their constructive comments.

{\small
\bibliographystyle{ieee_fullname}
\bibliography{egbib}
}

\end{document}